\documentclass[11pt,a4paper]{article}
\usepackage[hyperref]{emnlp-ijcnlp-2019}
\usepackage{times}
\usepackage{latexsym}

\usepackage{soul}
\usepackage{url}
\usepackage{amsfonts}
\usepackage{graphicx}
\usepackage{amsmath}
\usepackage{booktabs}
\usepackage[ruled]{algorithm2e}

\usepackage{amsmath}
\usepackage{booktabs}
\usepackage{color}
\usepackage{graphicx}
\usepackage{amsmath}
\usepackage{subfigure}
\usepackage[shortlabels]{enumitem}

\usepackage{url}

\usepackage{multirow}

\def\JM{{\mathcal J}}

\def\RM{{\mathcal R}}

\def\TM{{\mathcal T}}

\def\WM{{\mathcal W}}

\def\RB{{\mathbb R}}

\def\c{{\bf c}}

\def\e{{\bf e}}

\def\u{{\bf u}}

\def\v{{\bf v}}

\def\X{{\bf X}}

\def\0{{\bf 0}}
\def\1{{\bf 1}}

\usepackage{amsmath}

\aclfinalcopy % Uncomment this line for the final submission

\title{Multiplex Word Embeddings for Selectional Preference Acquisition}

\author{
Hongming Zhang$^\clubsuit$\thanks{~~Equal contribution.}, Jiaxin Bai$^\clubsuit$$^{*}$, Yan Song$^\spadesuit$, Kun Xu$^\heartsuit$,\\\textbf{Changlong Yu$^\clubsuit$, Yangqiu Song$^\clubsuit$, Wilfred Ng$^\clubsuit$, and Dong Yu$^\heartsuit$}\\
$^\clubsuit$Department of CSE, The Hong Kong University of Science and Technology\\
$^\spadesuit$Sinovation Ventures\\
$^\heartsuit$Tencent AI Lab\\
\{hzhangal, jbai, cyuaq, yqsong, wilfred\}@cse.ust.hk\\
clksong@gmail.com, \{kxkunxu, dyu\}@tencent.com\\
}

\date{}

\SetKwInput{KwInput}{Input}                % Set the Input
\SetKwInput{KwOutput}{Output}              % set the Output

\begin{document}
\maketitle
\begin{abstract}
Conventional word embeddings represent words with fixed vectors, which are usually trained based on co-occurrence patterns among words.
In doing so, however,
the power of such representations is limited, where the same word might be functionalized separately under different syntactic relations.
To address this limitation, one solution is to incorporate relational dependencies of different words into their embeddings.
Therefore,
in this paper, we propose a multiplex word embedding model, which can be easily extended according to various relations among words.
As a result,
each word has a center embedding to represent its overall semantics, and several relational embeddings to represent its relational dependencies.
Compared to existing models, our model can effectively distinguish words with respect to different relations without introducing unnecessary sparseness.
Moreover, to accommodate various relations, we use a small dimension for relational embeddings and our model is able to keep their effectiveness.
Experiments on selectional preference acquisition and word similarity demonstrate the effectiveness of the proposed model, and
a further study of scalability also proves that our embeddings only need 1/20 of the original embedding size to achieve better performance.
\end{abstract}

\section{Introduction}
\label{sec:introduction}

Representing words as distributed representations is an important way for machines to process lexical semantics, which attracts much attention in natural language processing (NLP) in the past few years \cite{DBLP:conf/nips/MikolovSCCD13,DBLP:conf/emnlp/PenningtonSM14,song-etal-2017-learning,song-etal-2018-directional,ijcai2018-607} with respect to its usefulness in many downstream tasks, e.g., parsing~\cite{chen2014fast}, machine translation~\cite{zou2013bilingual}, coreference resolution~\cite{lee2018higher}, etc.
Conventional word embeddings, e.g., word2vec~\cite{DBLP:conf/nips/MikolovSCCD13} and GloVe~\cite{DBLP:conf/emnlp/PenningtonSM14}, leverage the co-occurrence information among words to train a unified embedding for each word.
Such models are popular and the resulting embeddings are widely used owing to their effectiveness and simplicity.
However, these embeddings are not helpful for scenarios requiring words functionalizing separately under different situations,
where selectional preference (SP) \cite{wilks1975preferential} is a typical scenario.

In general, SP refers to that, given a word (predicate) and a dependency relation, human beings have certain preferences for the words (arguments) connecting to it.
Such preferences are usually carried in dependency syntactic relations, for example, the verb `sing' has plausible object words `song' or `rhythm' rather than other nouns such as `house' or `potato'.
With such characteristic, SP is proven to be important in natural language understanding for many cases and widely applied over a variety of NLP tasks, e.g., sense disambiguation \cite{resnik1997selectional}, semantic role classification \cite{semantic_role_classification},  and coreference resolution \cite{hobbs1978resolving,DBLP:conf/naacl/ZhangSS19,DBLP:conf/acl/ZhangSSY19}, etc.

\begin{figure}
    \centering
    \includegraphics[width=\linewidth, trim=0 20 0 50]{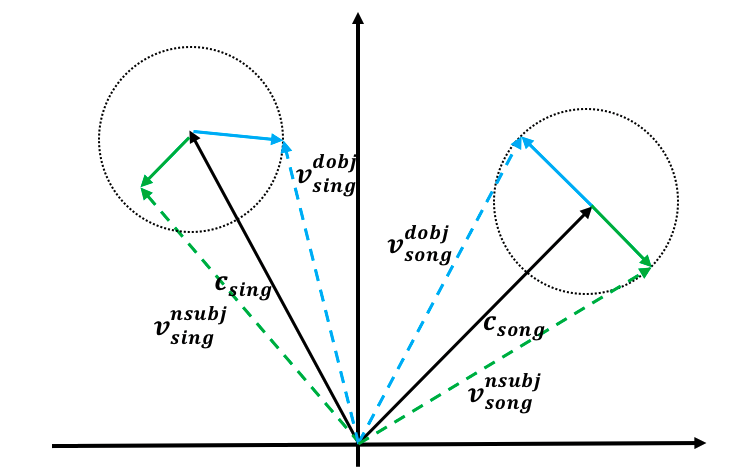}
    \caption{Illustration of the multiplex embeddings for `sing' and `song'.
    Black arrows present center embeddings for words' overall semantics; blue and green arrows refer to words' relational embeddings for relation-dependent semantics. 
    %\revisehm{Each word has a center embedding (black arrow) to represent its overall semantic, and several additional embeddings (blue and green arrows) to represent its relation-dependent semantics.
    All relational embeddings for each word are designed to near its center embedding.
    %All the separate embeddings are designed to be near the center embedding. 
    \emph{nsubj} and \emph{dobj} relations are used as examples.
    %}
    }
    % \vspace{-0.1in}
    \label{fig:demonstration}
\end{figure}

Conventional SP acquisition methods are either based on counting~\cite{resnik1997selectional} or complex neural network~\cite{DBLP:conf/emnlp/Cruys14}, and the SP knowledge acquired in either way can not be directly leveraged into downstream tasks.
On the other hand, the information captured by word embeddings can be seamlessly used in downstream tasks, which makes embedding a potential solution for the aforementioned problem.
However, conventional word embeddings using one unified embedding for each word are not able to distinguish different relations types (such as various syntactic relations, which is crucial for SP) among words.
For example, such embeddings treat `food' and `eat' as highly relevant words but never distinguish the function of `food' to be a subject or an object to `eat'.
To address this problem, the dependency-based embedding model \cite{DBLP:conf/acl/LevyG14} is proposed to treat a word through separate ones, e.g., `food@dobj' and `food@nsubj', under different syntactic relations, with the skip-gram~\cite{DBLP:conf/nips/MikolovSCCD13} model being used to train the final embeddings.
However, this method is limited in two aspects.
First, sparseness is introduced because each word is treated as two irrelevant ones (e.g., `food@dobj' and `food@nsubj'), so that the overall quality of learned embeddings is affected.
Second, the resulting embedding size is too large\footnote{Assuming one has 200,000 words in the vocabulary and 20 dependency relations, and follow conventional approaches \cite{DBLP:conf/nips/MikolovSCCD13,DBLP:conf/emnlp/PenningtonSM14} to set the embedding dimension to 300, the resulting embedding size will be about 10 Gigabytes.}, which is not appropriate either for storage or usage.

Therefore, in this paper,
we propose a multiplex word embedding (MWE) model, which can be easily extended to various relations between two words.
A multiplex network embedding model was originally proposed for modeling multiple relations among people in a social network~\cite{DBLP:conf/ijcai/ZhangQYS18}.
Interestingly, we found it also useful in 
%In this work, we extend it to the NLP domain, trying to
capturing various relations among different words.
One example is shown in Figure~\ref{fig:demonstration}. `sing' and `song' are highly related to each other with the `predicate-objective' rather than the 'predicate-subject' relation.
In our model, each word has a group of embeddings, including a center embedding representing its general semantics, and several embeddings representing their relation-dependent semantics.
To ensure that the embeddings of the same word (under different relations) are similar to each other, we limit the Euclidean norm of relation-dependent embeddings within a small range,
as shown in Figure~\ref{fig:demonstration}.
Moreover, considering that there could be many relations among words, if using a conventional dimension setting for embeddings to encode relations, the overall embedding size would be too big to be used in downstream tasks.
To deal with it, we propose to use a small dimension for relation-dependent embeddings and use a transformation matrix for each relation to project them into the same space of the center embeddings.
Thus the two types of embeddings can be jointly trained and the quality of the relation-dependent ones are guaranteed.

Experiments are conducted on SP acquisition over different dependency relations to evaluate whether the learned embeddings effectively capture words' semantics over these relations.
In addition, the word similarity measurement is used to assess how well words' general semantics are learned in our model.
Both evaluations confirm the superiority of our model, where the SP information is effectively preserved and the words' overall semantics are enhanced.
Particularly, further analysis also indicates that our MWE embeddings are more powerful than all existing embedding methods in SP acquisition with around 1/20 in their size comparing to 
previous embeddings \cite{DBLP:conf/acl/LevyG14}.
All code and resulting embeddings are available at: \url{https://github.com/HKUST-KnowComp/MWE}.

\section{Multiplex Word Embeddings}\label{sec:model}

\subsection{Model Overview}

As introduced in Section~\ref{sec:introduction}, encoding selectional preference information into embeddings can be conducted by modeling word association patterns under different dependency relations.
Similar to \cite{DBLP:conf/acl/LevyG14}, the proposed MWE model also distinguishes different relations among words and learns separate embeddings for them on each dependency edge.

Formally, let $\WM$ be the vocabulary set containing $n$ words $w_1, w_2, ..., w_n$ and $\RM$ the relations set containing $m$ relations $r_1, r_2, ..., r_m$,
the proposed model is expected to produce $m+1$ embeddings for each word, where there are $m$ relational embeddings representing relations and a center embedding $\c$ for the general semantics.
Particularly,
each relational embedding has an overall and a local version, denoted as $\v$ and $\u$, where $\u$ records the difference between $\v$ and $\c$.
For both the relational embeddings ($\v$ and $\u$) and the center embedding,
we use a similar way as that in word2vec \cite{DBLP:conf/nips/MikolovSCCD13} that each one
has two variants to represent the semantics of a word $w$ being the head or tail in different relations.
We denote the head and tail embeddings as $\u_{h,w}^r \in \RB^s$ and $\u_{t,w}^r \in \RB^s$ for the local embeddings of word $w$ under relation type $r$ and 
$\c_{w,h}  \in \RB^d$ and $\c_{w, t} \in {\RB}^d$ for the center embedding, respectively,
where $s$ is the dimension for $\u$ and $d$ the dimension for $\c$.

Although learning on similar information as that in \cite{DBLP:conf/acl/LevyG14}, 
to reduce the sparseness of introducing $m$ relational embeddings for each word,
we do not treat each word as multiple separate prototypes.
Instead, we use its center embedding to transfer information among different relations and
the sum of the center embedding and a local embedding to represent the final embedding for the corresponding relation\footnote{
Conceptually shown in Figure \ref{fig:demonstration}, to ensure the resulting final embeddings not too far away from the center embeddings, we add restriction to the Euclidean norm of the local relational embeddings.
}.
Moreover, considering there are various relations among words, we use a lower dimension $s$ for storage compression, with $s < d$.
Thus, a transformation matrix $\X$ is introduced to transform $\u$ into the same vector space of $\c$.
As a result, the final embeddings $\v$ for head and tail of $w$ under relation $r$ are formulated as:

\begin{equation}
\label{eq:decompose}
\begin{split}
\small
\v_{h,w}^r &= \c_{h,w} +  {\X_h^r}^T \u_{h,w}^r,\\ 
\v_{t,w}^r &= \c_{t,w} +  {\X_t^r}^T \u_{t,w}^r.
\end{split}
\end{equation}

\subsection{Learning the MWE model}

To train $\v_{h,w}^r$ and $\v_{t,w}^r$ for each $w$ under $r$,
we adopt the negative sampling strategy to conduct the learning process.
Specifically, for each $r$, we use a relation tuple set $\TM_r$, with each tuple $t = (w_h, r, w_t) \in \TM_r$, where $w_h$ is the head word and $w_t$ the tail word.
For each $t$, we randomly generate two negative tuples by replacing $w_h$ and $w_t$ with the randomly selected fake head $w_h^\prime$ and tail $w_t^\prime$ respectively.
Then the learning process is expected to distinguish the positive tuple against the negative ones.
Therefore, formally,
we maximize
\begin{equation}\label{eq:objective}
\small
\JM = \frac{1}{|\TM_r|} \sum_{t \in \TM_r}log\frac{e^{f(w_h,r,w_t)}}{e^{f(w_h,r,w_t)}+e^{f(w_h,r, w_t^\prime)}+e^{f(w_h^\prime,r, w_t)}},
\end{equation}
over all tuples in $\TM_r$ for each $r$,
with $f(\cdot)$ evaluating $w_h$ and $w_t$ being positive samples under $r$ through
\begin{equation}\label{eq:scoring_function}
f(w_h,r,w_t) = {\v_{h,w_h}^r}^\top \cdot \v_{t,w_t}^r.
\end{equation}

For each ($w_h$, $w_t$), we use cross entropy as the loss function
\begin{equation}
\small
\label{eq:loss}
\begin{split}
    E=&-\log \sigma ({\v_{h,w_h}^r}^\top \cdot \v_{t,w_t}^r) - \log \sigma (-{\v_{h,w_h}^r}^\top \cdot \v_{t,w_t^\prime}^r)\\
    & - \log \sigma (-{\v_{h,w_h^\prime}^r}^\top \cdot \v_{t,w_t}^r),
\end{split}
\end{equation}
to measure the learning effect for Eq. (\ref{eq:objective}),
with $\sigma$ denoting the sigmoid function.

Combined with Eq. (\ref{eq:decompose}), the training process is thus to update $c$, $u^r$, and $X^r$ with the gradients passed from the loss function using stochastic gradient descent (SGD).

In detail, take $w_h$ as an example, 
its center embedding $\c_{h,w_h}$ is updated by
\begin{equation}
\label{eq:c}
\begin{split}
\c_{h,w_h} &= \c_{h,w_h} - \lambda \cdot \eta \cdot \frac{\partial E}{\partial \v_{h,w_h}^r}\\ 
 &= \c_{h,w_h} - \lambda \cdot \eta \cdot e(w_h,w_t) \cdot \v_{t,w_t}^r,
\end{split}
\end{equation}
where 
$e(w_h,w_t)$ = $\sigma ({\e_{h,w_h}^r}^\top \cdot \e_{t,w_t}^r)-t_k$, with
$t_k = 1$ if ($w_h$, $w_t$) is positive sample and $t_k = 0$ for negative ones.
$\eta$ is the discounting learning rate.
$\lambda$ is an alternating weight to control the contribution of the gradient, ranging from 0 to 1.

Moreover, $X^r$ and $\u^r$ are updated as follows:
\begin{equation}
\small
\label{eq:u}
\begin{split}
\u_{h,w_h}^r &= \u_{h,w_h}^r - (1-\lambda) \cdot \eta \cdot \frac{\partial E}{\partial \v_{h,w_h}^r} \cdot \frac{\partial \X_h^r \u_{h,w_h}^r}{\partial \u_{h,w_h}^r}\\ 
 &= \u_{h,w_h}^r - (1-\lambda) \cdot \eta \cdot e(w_h,w_t) \cdot \X_h^r \cdot \v_{t,w_t}^r,
\end{split}
\end{equation}
% \vspace{-0.1in}
\begin{equation}
\small
\label{eq:X}
\begin{split}
\X_h^r &= \X_h^r - (1-\lambda) \cdot \eta \cdot \frac{\partial E}{\partial \v_{h,w_h}^r} \cdot \frac{\partial \X_h^r \u_{h,w_h}^r}{\partial \X_h^r}\\ 
 &= \X_h^r - (1-\lambda) \cdot \eta \cdot e(w_h,w_t) \cdot \u_{h,w_h}^r \cdot \v_{t,w_t}^r,
\end{split}
\end{equation}

Meanwhile, 
$\c_{t,w_t}$, $\u_{t,w_t}^r$, and $\X_t^r$ are updated in the same way of $\c_{h,w_h}$, $\u_{h,w_h}^r$, and $\X_h^r$ following
Eqs. (\ref{eq:c}), (\ref{eq:u}), and (\ref{eq:X}).

% \reviseyq{I still think the equations above are redundant and even confusing. Do we really need $\c_{t,w}$? Can we just say $\c_{w}$? Then we have $\c_{w_h}$ and $\c_{w_t}$ to distinguish head and tail, while still maining only ONE center embedding for each word.}

\begin{algorithm}[t]
\SetAlgoLined

 \KwInput{Relation specific tuple sets $\TM_1,\ldots,\TM_m$, $d$, $s$, $a$, and $\eta$.} 
  \textbf{Output:} $\c_w$, $\u_w^r$, and $\X^r$.\\
 Initialize $\c_w$, $\u_w^i$ and $\X^i$ randomly.\\ 
 \For{Each Iteration k}{
    Update $\lambda$.\\
    \For{Each tuple $t_p$ = ($w_h$, $r$, $w_t$) $\in \TM_r$}{
        Randomly generate the two negative examples $t_{n1}$ = ($w_h^\prime$, $r$, $w_t$) and $t_{n2}$ = ($w_h$, $r$, $w_t^\prime$).\\
        $\TM_{tmp} = [t_p,t_{n1},t_{n2}]$ \\
        \For{$t \in \TM_{tmp}$}{
            $e(w_h, w_t) = \sigma ({\textbf{v}_{h, w_h}^{\prime}}^T \cdot \textbf{v}^i_{t, n_t})-t_k$ \\
            Update $\c_{h,w_h}$, $\u_{h,w_h}^r$, $\X_h^r$, $\c_{t,w_t}$, $\u_{t,w_t}^r$, and $\X_t^r$ based on Eqs.~(\ref{eq:c}), \ref{eq:u}), and (\ref{eq:X}).\\
            \If{$\| {\X_h^r}^T \u_{h, w_h}^r \| > a$ }{
                Update $\X_h^r$ and $\u_{h, w_h}^r$.
            }
            \If{$\| {\X_t^r}^T \u_{t, w_t}^r \| > a$}{
                Update $\X_t^r$ and $\u_{t, w_t}^r$.
            }
        }
    }
 }

 \caption{Multiplex Word Embedding}\label{algorithm}
\end{algorithm}

The above learning process is conducted on all $m$ relations at the same time.
During the training, the Euclidean norm of $\X^T \u $ for all embeddings is constrained to have a maximum semantic drifting range $a$, whose value controls the distance between the local relational embeddings and the center embedding.
Specifically, if $\| \X^T \u \|$ equals to $a^\prime$, which is greater than $a$, we scale down $\X^T$ and $\u$ with $\sqrt{a^\prime \cdot \frac{1}{ka}}$, where $k$ is a scaling parameter set to 0.8 throughout this work.

The learning processing is summarized in Algorithm~\ref{algorithm}.
For complexity analysis, it is obvious drawn that $m$ relation types and $n$ words result in
%Assuming that we have $M$ relation types and $| \NM |$ words, the time complexity of our model is $O(M| \NM |)$. The memory complexity of our model is $O((d+s*M)| \NM |)$.
$O(mn)$ and $O((d+s*m)*n)$ for time and space complexities, respectively.

\subsection{The Alternating Optimization Strategy}

To balance the learning process between the overall semantics and the relation-specific information, we adopt an alternating optimization strategy\footnote{Alternating optimization was originally proposed to train different parameters in a sequential manner and applied in various areas.} \cite{bezdek2003convergence} to adjust $\lambda$ based on different stage of the training instead of using a fixed weight for $\lambda$.
Specifically, considering $\c$ is more reliable and possesses more information while $\u$ is learned with shared $\c$, we alternatively update $\c$ and $\u$ upon the convergence of $\c$.
As a result,
%in model training,
%we focus more on the center embeddings at the beginning stage and then start updating \textcolor{green}{shift embeddings}.
%As a result,
we set $\lambda$ to 1 in the first half of the training process and 0 afterwards.

\section{Experiments}\label{sec:pseudo_disambiguation}

%In this section, we conduct experiments to demonstrate the proposed multi-relational word embedding can better represent the semantic information of words over various relations. We first introduce the corpus used to train the embeddings and baseline models. After that, we conduct experiment on two tasks: selectional preference acquisition and word similarity.
Experiments are conducted to evaluate how our embeddings are performed on SP acquisition and word similarity measurement.

\subsection{Implementation Details}
We use the English Wikipedia\footnote{https://dumps.wikipedia.org/enwiki/} as the training corpus.
The Stanford parser\footnote{https://stanfordnlp.github.io/CoreNLP/} is used to obtain dependency relations among words. 
For the fair comparison, we follow existing work and set
$d = 300$, $s = 10$, and $a = 1$.
Following \cite{DBLP:journals/coling/KellerL03} and \cite{DBLP:conf/emnlp/Cruys14}, we select three dependency relations (\emph{nsubj}, \emph{dobj}, and \emph{amod})
as follows:

\begin{itemize}[leftmargin=*]
    \item \emph{nsubj}: The preference of subject for a given verb.
    For example, it is plausible to say `dog barks' rather than `stone barks'.
    The verb is viewed as the predicate (head) while the subject as the argument (tail).
    \item \emph{dobj}: The preference of object for a given verb.
    For example, it is plausible for `eat food' rather than `eat house'.
    The verb is viewed as the predicate (head) while the object as the argument (tail).
    \item \emph{amod}: The preference of modifier for a given noun.
    For example, it is plausible to say `fresh air' rather than `solid air'.
    The noun is viewed as the predicate (head) while the adjective as the argument (tail).
\end{itemize}

\subsection{Baselines}
We first compare the proposed multiplex embedding with the following embedding models. As it is trivial to apply these embedding models in downstream tasks, we label these models as downstream friendly.

\begin{table*}[t]
    \centering
    % \small
    \begin{tabular}{l|c||ccc||cccc}
    \toprule
            \multirow{2}{*}{Model}          & \multirow{2}{*}{Downstream} 
            & \multicolumn{3}{c||}{Keller}  & \multicolumn{4}{c}{SP-10K} \\
            \cline{3-9}
           & & dobj & amod & average & nsubj & dobj & amod & average   \\
            
    \midrule
    word2vec  & Friendly & 0.29 & 0.28 & 0.29 & 0.32 & 0.53 & 0.62 & 0.49\\
    GloVe  & Friendly & 0.37 & 0.32 & 0.35 & 0.57 & 0.60 & 0.68 & 0.62\\
    % \midrule
    D-embeddings  & Friendly & 0.19 & 0.22 & 0.21 & 0.66 & 0.71 & 0.77 & 0.71 \\
    \midrule
    ELMo  & Friendly & 0.23 & 0.06 & 0.15 & 0.09 & 0.29 & 0.38 & 0.25\\
    BERT (static) & Friendly & 0.11 & 0.05 & 0.08 & 0.25 & 0.32 & 0.27 & 0.28\\
    BERT (dynamic) & Friendly & 0.19 & 0.23 & 0.21 & 0.35 & 0.45 & 0.51 & 0.41\\
    \midrule
    PP & Unfriendly & \textbf{0.66} & 0.26 & 0.46 & 0.75 & 0.74 & 0.75 & 0.75\\
    DS & Unfriendly & 0.53 & 0.32 & 0.43 & 0.59 & 0.65 & 0.67 &  0.64 \\
    NN & Unfriendly & 0.16 & 0.13 & 0.15 & 0.70 & 0.68 & 0.68 & 0.69\\

    \midrule
    % MWE ($c$) &  0.71 & \textbf{0.79} & \textbf{0.78} & 0.76 \\
    MWE & Friendly & 0.63 & \textbf{0.43}$^\dag$ & \textbf{0.53}$^\dag$ & \textbf{0.76} & \textbf{0.79}$^\dag$ & \textbf{0.78} & \textbf{0.78}$^\dag$\\
    \bottomrule
    \end{tabular}
    \caption{
    \small
    %Experiment Result on selectional preference acquisition over three different relations are reported.
    Results on different SP acquisition evaluation sets.
    As Keller is created based on the PP distribution and has relatively small size while SP-10K is created based on random sampling and has a much larger size, we treat the performance on SP-10K as the main evaluation metric.
    Spearman's correlation between predicated plausibility and annotations are reported.
    The best performing models are denoted with bold font.
    $\dag$ indicates statistical significant (p \textless 0.005) overall baseline methods.
    }
    \label{tab:main_result}
\end{table*}

\begin{itemize}[leftmargin=*]
\item
\textbf{word2vec}, the embedding model proposed by \citeauthor{DBLP:conf/nips/MikolovSCCD13} \shortcite{DBLP:conf/nips/MikolovSCCD13}.
We use the skip-gram model for this baseline.
\item
\textbf{GloVe} \cite{DBLP:conf/emnlp/PenningtonSM14}, learning word embeddings by matrix decomposition on word co-occurrences.
\item
\textbf{D-embeddings}, the model proposed by \citeauthor{DBLP:conf/acl/LevyG14} \shortcite{DBLP:conf/acl/LevyG14} uses a skip-gram framework to encode dependencies into embeddings with multi-prototypes of words.
\end{itemize}
To investigate whether the SP knowledge can be captured by the pretrained contextualized word embedding models, we also treat following pre-trained models as baselines.
\begin{itemize}[leftmargin=*]
\item
\textbf{ELMo} \cite{DBLP:conf/naacl/PetersNIGCLZ18}, a pretrained language model with contextual awareness. We use its static representations of words as the word embedding.
\item
\textbf{BERT} \cite{devlin2018bert}, a pretrained bi-directional contextualized word embedding model with state-of-the-art performance on many NLP tasks.
%We use the static representation of words as the word embedding.
% For the ELMo and BERT, 
\end{itemize}

Besides those embedding methods, we also compare with following conventional SP acquisition methods to demonstrate the effectiveness of the proposed multiplex embedding model, as it is still unclear how to leverage these methods in downstream tasks, we label these methods as downstream unfriendly.
\begin{itemize}[leftmargin=*]
    \item 
    \textbf{Posterior Probability (PP)}~\cite{resnik1997selectional}, a counting based method for the selectional preference acquisition task.
    \item
    \textbf{Distributional Similarity (DS)}~\cite{DBLP:journals/coling/ErkPP10}, a method that uses the similarity of the embedding of the target argument and average embedding of observed golden arguments in the corpus to predict the preference strength. 
    \item
    \textbf{Neural Network (NN)}~\cite{DBLP:conf/emnlp/Cruys14}, an NN-based method for the SP acquisition task. This model achieves the state-of-the-art performance on the pseudo-disambiguation task.
\end{itemize}

\begin{table}[t]
% \footnotesize
\small
\centering
\begin{tabular}{c|ccc}
	\toprule
    SP Evaluation Set  & \#W & \#P  \\
    \midrule
    Keller~\cite{DBLP:journals/coling/KellerL03} & 571   & 360   \\
    % Pado~\cite{pado2006combining}    & 180   & 207   \\
    SP-10K~\cite{DBLP:conf/acl/ZhangDS19}    & 2,500 & 6,000  \\
    \bottomrule
	\end{tabular}
	\caption{Statistics of Human-labeled SP Evaluation Sets. \#W and \#P indicate the numbers of words and pairs, respectively. As different datasets have different SP relations, we only report statistics about `nsubj', `dobj', and `amod' (if available).}
% 	\vspace{-0.15in}
	\label{tab:stat}

\end{table}

For word2Vec and GloVe, we use their released code. 
For D-embedding,
we follow their original paper using the Gensim package\footnote{https://radimrehurek.com/gensim/models/word2vec.html}.
The dimensions of the aforementioned embeddings are set to 300 according to their original settings.
For ELMo and BERT, we use their pre-trained models.
As BERT was not originally designed for word level semantics tasks, for the selectional preference acquisition task, we compare with two variations of the original BERT model: (1) BERT (static), which extracts static embedding from the BERT model in a similar way as that from the ELMo model; (2) BERT (dynamic), which takes a pair of words as input and produces a plausibility score for that pairs of words. The main difference between the static and dynamic version of BERT is that in the dynamic version, the contextual information can be fully utilized, which is more similar to the training objective of BERT.

\begin{table}[t]
    \centering
    \small
    \begin{tabular}{c|c||c|c|c|c}
    \toprule
         
         \multicolumn{2}{c||}{Model}   & ~~noun~~ & ~~verb~~ & adjective  & overall \\
            \midrule
    
    \multicolumn{2}{c||}{word2vec}   & 0.41 & 0.28  & 0.44 & 0.38 \\
    \multicolumn{2}{c||}{Glove}  & 0.40 & 0.22 & 0.53 & 0.37\\
    \midrule
    \multicolumn{2}{c||}{D-embedding}   & 0.41 & 0.27 & 0.38 &  0.36 \\
    % \midrule
    % \multicolumn{2}{c||}{ELMo}   & 0.438 (-) & 0.382 (-) & 0.551 (-)  & 0.434 (-) \\  
    % \multicolumn{2}{c||}{BERT}   & 0.531 (-) & 0.342 (-) & 0.532 (-) & 0.486 (-) \\
    \midrule
    \multirow{4}{*}{nsubj} & h & 0.46  & 0.29  & 0.54 & 0.43\\
    & t &  0.45 & 0.25 & 0.48 & 0.40 \\
    & h+t & 0.44  & 0.23 & 0.50 & 0.40 \\
    & [h,t] & 0.47 & 0.27 & 0.51 & 0.42 \\
    \midrule
    \multirow{4}{*}{dobj} & h & 0.46 & 0.27 & 0.45 & 0.41 \\
    & t & 0.45 & 0.23 & 0.46 & 0.40 \\
    & h+t & 0.45 & 0.20 & 0.45 & 0.38 \\
    & [h,t] & 0.46 & 0.25 & 0.48 & 0.42 \\
    \midrule
    \multirow{4}{*}{amod} & h & 0.47 & 0.25 & 0.52 & 0.37 \\
    & t & 0.46 & 0.24 & 0.50 & 0.38 \\
    & h+t & 0.46 & 0.24 & 0.52 & 0.38 \\
    & [h,t] & 0.47 & 0.26 & 0.52 & 0.38 \\
    % \midrule
    % \multirow{4}{*}{hy-hy} & h & xxx (xxx) & xxx (xxx) & xxx (xxx) & xxx (xxx) \\
    % & t & xxx (xxx) & xxx (xxx) & xxx (xxx) & xxx (xxx) \\
    % & h+t & xxx (xxx) & xxx (xxx) & xxx (xxx) & xxx (xxx) \\
    % & [h,t] & xxx (xxx) & xxx (xxx) & xxx (xxx) & xxx (xxx) \\
    \midrule
    \multirow{4}{*}{center} & h & 0.51 & \textbf{0.33} & \textbf{0.57} & \textbf{0.48} \\
    & t & 0.51 & 0.30 & 0.56 & 0.47 \\
    & h+t & \textbf{0.52} & 0.31 & 0.54 & 0.46 \\
    & [h,t] & 0.51 & 0.32 & 0.57 & \textbf{0.48} \\
    \bottomrule
    \end{tabular}
    \caption{
    \small
    Spearman's correlation of different embeddings for the WS measurement. 
    `nsubj', `dobj', `amod' represents the embeddings of the corresponding relation and `center' indicates the center embeddings.
    h, t, h+t, and [h,t] refer to the head, tail, sum of two embeddings, and the concatenation of them, respectively.
    The best scores are marked in bold fonts.}
    \label{tab:WSResult}
\end{table}

\subsection{Selectional Preference Acquisition}

The first task in our experiment is SP acquisition.
Currently, there are two methods that we can use to evaluate the quality of extracted SP knowledge: Pseudo-disambiguation~\cite{DBLP:conf/acl/RitterME10} and human labeled datasets~\cite{mcrae1998modeling,DBLP:journals/coling/KellerL03}.
However, as shown in ~\cite{DBLP:conf/acl/ZhangDS19}, pseudo-disambiguation selects positive SP tuples from the corpus and generate negative SP tuples by randomly replacing the head/tail.
As a result, pseudo-disambiguation only evaluates how good the model fits the corpus rather than evaluating the SP acquisition models based on ground truth.
Thus, in this section, we evaluate different SP acquisition methods with ground truth (human labeled datasets).
Two representative datasets, Keller~\cite{DBLP:journals/coling/KellerL03} and SP-10K~\cite{DBLP:conf/acl/ZhangDS19}, are selected as the benchmark datasets.

In each dataset, for each SP relation, various word pairs are provided.
For each of the word pairs, the datasets also provide the annotated plausibility score of how likely a preference exists between that word pair under the corresponding SP relation.
Thus the job for all the models is to predict the plausibility score for all word pairs under different SP relation settings. 
After that, we follow the conventional setting that uses the Spearman's correlation to assess the correspondence between the predicted plausibility scores and human annotations on all word pairs for all SP relations.
The statistics of these datasets are shown in Table~\ref{tab:stat}.

For embedding based methods (word2vec, GloVe, D-embedding\footnote{For D-embedding, since it provides embeddings for different relations (e.g., `food@nsubj', `food@dobj), we follow their work and directly use
%their original setting and compute the similarity of
embeddings over certain relations to predict the score for the tested pairs. 
For example, given the test tuple (`eat', \emph{dobj}, `food'), we compute the cosine similarity of `eat' and `food@dobj' rather than `food@subj'.}, and MWE\footnote{For the proposed model MWE, we use the cosine similarity between $v_{h,w_h}$ and $v_{t, w_t}$ as the predicted plausibility.}), we follow previous work \cite{DBLP:conf/nips/MikolovSCCD13,DBLP:conf/acl/LevyG14} and use the cosine similarity between embeddings of head and tail words to predict their relations. 
For conventional SP acquisition methods (PP, DS, and NN), we follow their original paper to compute the plausibility scores. 

\begin{table}[t]
    \centering
    \small
    \begin{tabular}{c||c|c|c}
    \toprule
      Model  &  WS & Dimension & Training Time\\
      \midrule 
      ELMo  & 0.434 & 512 &  $\approx$40 \\
      BERT & 0.486 & 768 & $\approx$300 \\
        \midrule
        MWE & 0.476 & 300 & 4.17 \\
         \bottomrule
    \end{tabular}
    \caption{
    \small
    Comparison of MWE against language models on the WS task. Overall performance, embedding dimension, and training time (days) on a single GPU are reported.}
    \label{tab:ws-lm}
\end{table}

The experimental results are shown in Table~\ref{tab:main_result}.
As Keller is created based on the PP distribution and have relatively small size while SP-10K is created based on random sampling and has a much larger size, we treat the performance on SP-10K as the major evaluation.
Our embeddings significantly outperform other baselines, especially embedding based baselines.
The only exception is PP on the Keller dataset due to its biased distribution.
In addition, there are other interesting observations.
First, compared with `dobj' and `nsubj', `amod' is simpler for word2vec and GloVe. The reason behind is that conventional embeddings only capture the co-occurrence information, which is enough to predict the selectional preference of `amod' rather than `nsubj' or `dobj'\footnote{The only possible SP relation between nouns and adjectives is `amod', while multiple SP relations could exist between nouns and verbs, and co\_occurrence information cannot effectively distinguish them.}. 
Second, even though large-scale contextualized word embedding models like ELMo and BERT have been proved useful in many other tasks, they are still limited in learning specific and detailed semantics and thus perform inferior to our model in the SP acquisition task.
For Example, ELMo and BERT can know that there is a strong semantic connection between `eat' and `food', but they do not know whether `food' is a plausible subject or object of `eat'.
Third, surprisingly, although NN achieves the state-of-the-art performance on the pseudo-disambiguation task, its performance is not satisfying against human annotation, especially on Keller, which is probably because the NN model overfits the training data, whose distribution is different from human SP knowledge.

\subsection{Word Similarity Measurement}

In addition to SP acquisition, we also evaluate our embeddings on word similarity (WS) measurement
to test whether the learned embedding can effectively capture the overall semantics.
%information, we also conduct experiments on the general word similarity task.
We use
SimLex-999 \cite{DBLP:journals/coling/HillRK15} as the evaluation dataset for
%word similarity measurement,
this task because it contains different word types, i.e.,
%which contains
666 noun pairs, 222 verb pairs, and 111 adjective pairs.
We follow the conventional setting that uses the Spearman's correlation to assess the correspondence between the similarity scores and human annotations on all word pairs.
%The experimental results of all combinations are reported.
Evaluations are conducted on the final embeddings $\v$ for each relation and the center ones.
% , with results reported in Table \ref{tab:WSResult}.

\begin{figure}[t]
    \centering
    \includegraphics[width=0.8\linewidth, trim=0 0 0 10]{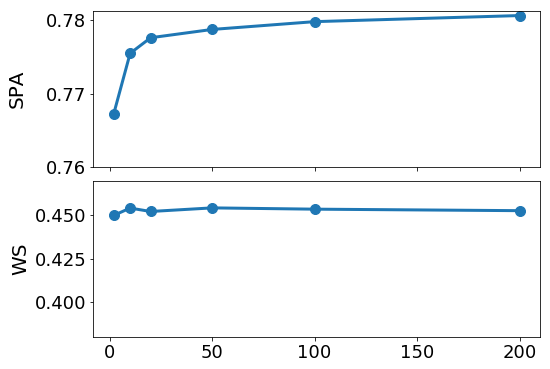}
    \caption{Effect of the different $s$ on SP acquisition (SP-10K) and WS tasks.
    %$s$. Performance on SPA and WS are reported. \revisehm{All the other settings are unchanged.}
    }
    \label{fig:dimension}
\end{figure}

%\revisehm{We first compare our model with all the embedding models in Table \ref{tab:WSResult}}
%As shown in table~\ref{tab:WSResult},
Results are reported in Table \ref{tab:WSResult}
with
%and there are
several observations.
First,
%our model beats all the other models significantly over nouns.
our model achieves the best overall performance and significantly better on nouns,
%Thanks to that, our model also achieves the best overall performance.
which can be explained by that nouns appear in all three relations while most of the verbs and adjectives only appear in one or two relations.
This result is promising since it is
analyzed by \citeauthor{solovyev2017dynamics} \shortcite{solovyev2017dynamics} that two-thirds of the frequent words are nouns;
thus there are potential benefits if our embeddings are used in downstream NLP tasks.
Second, the center embeddings achieve the best performance against all the other relation-dependent embeddings, which
demonstrates the effectiveness of our model in learning relation-dependent information over words and also enhancing their overall semantics.

We also compare MWE with pre-trained contextualized word embedding models in Table \ref{tab:ws-lm} for this task, with
overall performance, embedding dimensions, and training times reported.
%and compared with ours.
It is observed that
%From the result we can see
that MWE outperforms ELMo and achieves comparable results with BERT with smaller embedding dimension and much less training complexities.

% \begin{table*}[]
%     \centering
%     \small
%     \begin{tabular}{l||c|c|c|c||c|c|c|c}
%     \toprule
%          \multirow{2}{*}{Model} & \multicolumn{4}{c||}{Pseudo-disambiguation} & \multicolumn{4}{c}{Word Similarity}\\
%         %  \midrule
%         \cline{2-5} \cline{6-9}
%                                 & nsubj & dobj & amod & overall & ~~noun~~ & ~~verb~~ & adjective  & Overall \\
%                                 \midrule
    
%     word2vec & 70.36 (0.04) & 77.70 (0.06) & 75.61 (0.04) & 74.56 (0.05)   & 0.41 & 0.27& 0.45  & 0.38  \\
%     GloVe & 90.78 (0.04) & 91.55 (0.06) & 86.04 (0.08) & 89.39 (0.05)   & 0.40 & 0.15& 0.59  & 0.37 \\
%     \midrule
%     ELMo$^\star$ & 57.09 (0.08) & 64.21 (0.05) & 54.51 (0.07) & 58.60 (0.07)   & 0.44 & 0.38 & 0.55  & 0.43  \\                
%     \midrule
%     D-embedding & 86.14 (0.05) & 81.97 (0.04) & 87.03 (0.08) & 85.05 (0.06)   & 0.33 & 0.20 & 0.28  & 0.30  \\
%     \midrule
%     Our Model & \textbf{?} (?) & \textbf{?} (?) & \textbf{?} (?) & \textbf{?} (?)  & ? &  ? &  ? & ? \\
%     \bottomrule
%     \end{tabular}
%     \caption{Experiment Result on PD and WS, where accuracy is reported for PD with standard deviations in brackets and
%     Spearman correlation $\rho$ for WS.
%     $\star$ indicates pretrain model which is trained on a much larger corpus.}
%     \label{tab:Result}
%     \vskip -1em
% \end{table*}

\begin{figure}[t]
    \centering
    \includegraphics[width=0.8\linewidth, trim=0 0 0 10]{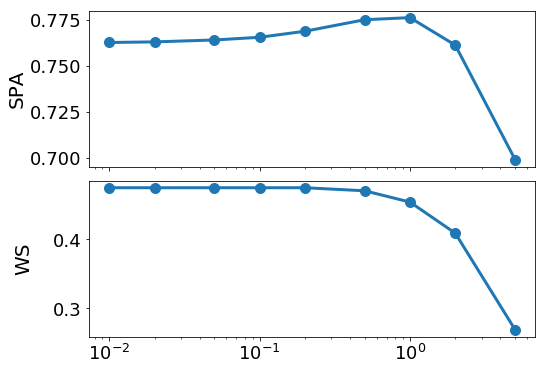}
    \caption{Effect of different $a$ on SP acquisition (SP-10K) and WS tasks.
    %. Performance on SPA and WS are reported. \revisehm{All the other settings are unchanged.}
    }
    \label{fig:limitation}
\end{figure}

\section{Analysis}
With the experiments on SP acquisition and word similarity measurement, we conduct further analysis for different settings for hyper-parameters and learning strategies, as well as the scalability of our model.
% Details are illustrated in the following subsections.

\subsection{Effect of the Local Embedding Dimension}

%In this section, we analyze the influence of $s$, which is the small embedding dimension.
We investigating the performance of MWE on SPA (SP acquisition) and WS tasks with different $s$.
As shown in Figure~\ref{fig:dimension}, the performance of our model is in an increasing trend when we increase the value of $s$.
%dimension of the additional vector.
When the dimension reaches 10, the performance almost reaches the top,
%reaches the top,
which confirms that the local relational embeddings can be effective in small dimensions (about 1/30 of the center dimension) when using center embeddings as the constraint.
%proves that with the help of center embedding, we can use a very low dimension (about 1/30 of the center dimension) vector to represent the relation-dependent information.

\subsection{Effect of the Semantic Drifting Range}

Similar to $s$, we also investigate the model performance with the influence of different constraint $a$.
It is observed in Figure \ref{fig:limitation} that, the constraint gets looser
with the increase of $a$.
For SP acquisition, in a small range, the model performance gradually improves along with the increasing value of $a$ because these embeddings are more flexible to capture intense relation-dependent information.
However, once the range passes a threshold (1 in our experiment), the embeddings get farther away from their general semantics and start to overfit, which is also observed in the WS experiment, and thus the performance drops monotonically.

\subsection{Effect of Alternating Optimization}
\label{sec:restriction}

\begin{table}[t]
    \centering
    \small
    \begin{tabular}{l||c|c}
    \toprule
            Training Strategy            & Averaged SPA & Overall WS  \\
    \midrule
    $\lambda$ = 1 & 0.762 & 0.476 \\
    $\lambda$ = 0 & 0.073 & 0.018  \\
    $\lambda$ = 0.5 & 0.493 & 0.323  \\
    \midrule
    Alternating optimization & 0.775 & 0.476 \\
    \bottomrule
    \end{tabular}
    \caption{
    \small
    Comparisons of different training strategies.
    %Effect of the alternating optimization strategy.
    % The results for averaged SP acquisition (SP-10K) and overall WS are computed from embeddings following the best settings in Table \ref{tab:main_result} and \ref{tab:WSResult}.
    %The averaged PD accuracy is calculated based on our final embeddings and the word similarity is calculated on SimLex-999 base on the best performed combination in Table \ref{tab:WSResult},
    %which is head `nsubj' embedding of `All relational' strategy and head center embeddings for all the others.
    % head center embeddings as it performs the best in previous experiments.
    }
    \label{tab:alternating}
    %\vskip -0.5em
\end{table}

An analysis is also conducted to demonstrate the effectiveness of the alternating optimization strategy.
As shown in Table \ref{tab:alternating}, we compare our model with several different strategies.
The first one is to put all weights to the center embedding (fix $\lambda$ to 1), which never updates the local relational embeddings.
As a result, it can achieve similar performance on word similarity measurement but is inferior in SP acquisition because no relation-dependent information is preserved.
The second strategy is to put all weights to the local relational embeddings (fix $\lambda$ to 0).
In this case, it performs very poor owing to the loss of general semantics.
%Last but not least, we also compare our model with the balance model, which fix $\lambda$ to 0.5.
Last but not least, the alternating optimization also outperforms the setting with a fixed $\lambda = 0.5$,
%For the experimental result, our model outperform the balanced model significantly on both tasks.
% The alternating optimization illustrates its superiority with the observations drawn from Table \ref{tab:alternating},
which can be explained by alternating optimization can first get a good overall semantic representation and then acquire the SP knowledge on top of that.
%The reason behind is that a good center embedding is very important for learning other relation-dependent embeddings.
To summarize, the alternating optimization provides an effective solution to training our model with a smart process in adjusting the contribution of different embeddings as well as
stabilizing their optimization.
%Thus compared with training all the embedding at the same time, focusing more on the center embedding first and then focus more on the other embeddings can help the whole training process become more stable.  

% \section{Visualization}

% \begin{figure}
%     \centering
%     \includegraphics[width=0.6\linewidth]{image/Visulization.png}
%     \caption{Visualization of selected resulted embedding. 
%     Other embeddings are hided for the clear representation. Closer distance indicates that two words are more similar and related in the embedding space.}
%     \label{fig:vis}
% \end{figure}

% In Figure~\ref{fig:vis}, we visualize the sampled word embeddings with t-SNE. The closer distance between embeddings indicates they are more similar in the embedding space and more related to each other.
% From the result, we can see that the distance between $sing\_dobj$ and $song\_dobj$ is smaller than $sing\_nsubj$ and $song\_nsubj$, which demonstrates that the learned embedding prefers `song' to be the object rather than the `subject' of `sing'
% Besides that, if we compare the distance between $sing\_dobj$ and $song\_dobj$ against $house\_dobj$, we can see that our model prefers 'song' rather than 'house' to be the object of `sing'.
% Last but not least, we can see that the separate embeddings of same words are very close, which is because we use the base embedding as the constraint during the training process.

\subsection{Scalability}

Scalability of embeddings is an important factor in real applications.
Practically, GPU or similar computation support is required to accommodate embeddings to applying neural models.
Normally, the current most affordable GPUs typically have a memory size limitation around 10 GB.
Thus, to ensure the learned embeddings can be used in downstream tasks, their size becomes an important concern when evaluating different embedding models, especially when there exist many different relations between words in our model.

\begin{figure}[t]
    \centering
    \includegraphics[width=0.85\linewidth, trim=30 15 0 20]{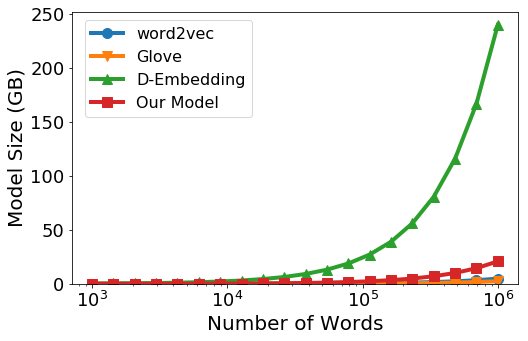}
    \caption{Scalability on embedding size against number of words.
    %All embedding dimensions are chosen based on their best performance.
    }
    \label{fig:Scalability}
\end{figure}

Assuming that there are 20 relations among words, the size of all embedding models with the increasing word numbers in Figure \ref{fig:Scalability}.
word2Vec and GloVe have the smallest size because they only train one embedding for each word while sacrificing the relation-dependent information among them.
As a comparison, D-embeddings are trained on multiple prototypes for each word to record relation information, thus their size dramatically explodes with the increasing of the vocabulary.
Consequently, it is easily computed for D-embeddings that 200,000 words will result in a 10GB model,
which is unfeasible to be used in downstream tasks.
Effectively with the small dimension for our local relational embeddings,
relation information can be preserved in a small-sized model, which shows a compatible space requirement with the conventional embeddings.

\section{Related Work}\label{sec:related_work}
Learning word embeddings has become an important research topic in NLP \cite{bengio2003-JMLR,turney2010-JAIR,collobert2011-JMLR1,ijcai2018-607}, with the capability of embeddings demonstrated in different languages \cite{ijcai2018-0608} and tasks such as parsing \cite{chen2014fast}, machine translation~\cite{zou2013bilingual}, coreference resolution~\cite{lee2018higher}, etc.
Conventional word embeddings \cite{DBLP:conf/nips/MikolovSCCD13,DBLP:conf/emnlp/PenningtonSM14} often leverage word co-occurrence patterns,
%to get the distributed representations of words.
resulting in a major limitation that they coalesce different relationships between words 
%such as SP over different relations or synonymy relations,
into a single vector space.
To address this limitation, dependency-based embedding model \cite{DBLP:conf/acl/LevyG14} was proposed to represent each word with several separate embeddings, and then
%This method
suffers from its sparseness and the huge size of the resulting embeddings.
%of the resulted embeddings.
%Different from them,
Alternatively,
%we propose a multi-relational embedding model, which contains one center embedding acting as the bridge between different relations and multiple-relation-dependent embedding to record the relation-dependent semantics.
our MWE model uses a set of (constrained and small) embeddings for each word to encode its general and relation-dependent semantics
%, which is constrained by the overall semantics, 
to resolve the sparseness and the size problem.
% , with a small dimension on relation embeddings so as to overcome the data sparseness. %problem.
%Moreover, the proposed conversion matrix can effectively shrink the resulted embedding size to 1/10 without influencing the overall performance.

Another line of work related to this paper is the research on SP, which is considered important in language understanding 
%As one important language phenomenon, Selectional Preference over dependency relations is considered to be related to semantics~\cite{wilks1975preferential}
and has been proved helpful in various downstream tasks \cite{DBLP:conf/coling/TangXZG16,resnik1997selectional,hobbs1978resolving,DBLP:conf/coling/InoueMOOI16,semantic_role_classification}.
Several studies attempted to acquire SP automatically from raw corpora \cite{resnik1997selectional,rooth1999inducing,DBLP:journals/coling/ErkPP10,santus2017measuring}.
However, they are designed specifically for SP acquisition and are limited in leveraging SP knowledge for downstream tasks.
Compared to them, the learned embeddings from our model are useful resources that incorporate SP knowledge and can be seamlessly leveraged in other NLP models.

% Detection of hypernymy, also known as lexical entailment, is viewed as an essential NLP task due to the important role it plays in a large number of tasks, such as taxonomy creation~\cite{snow2006semantic,navigli2011graph} and entailment prediction~\cite{dagan2013recognizing}.
% Conventionally, people design rules~\cite{hearst1992automatic,snow2005learning} to extract hypernymy-hyponymy pairs  from textual corpus, but the extracted knowledge cannot be easily used in downstream tasks like word embeddings.
% After that, supervised model~\cite{N18-1103} was proposed to inject hypernymy knowledge to the pretrained word embedding.
% However, as introduced by~\cite{shwartz2016hypernyms}, supervised methods suffer from the biase of limited training instances and noisy extracted knowledge.
% Different from them, in this paper, we leverage the unsupervised training to learn high-quality word embeddings to capture hypernymy knowledge, which are more robust.

\section{Conclusion}\label{sec:conclusion}

In this paper, we present a multiplex word embedding model encoding selectional preference information from word dependency relations.
Different from conventional embeddings, for each word, we proposed to use a set of embeddings to represent its general (center embedding) and relation-dependent (local relational embedding) semantics, where their combination present the final embedding for each word under particular relations.
%base embedding to represent its general semantics information and use multiple separate embeddings to preserve its semantics information over different relations.
Experiments on SP acquisition and word similarity measurement illustrated that our model better encodes SP knowledge than different baselines without harming its capability in representing general semantics.
Analysis is also conducted with respect to different settings and optimization strategies, as well as the effect of model size,
which further illustrates the validity and effectiveness of the proposed model and its learning process.
%In the future, we will explore how can we help downstream NLP tasks with SP knowledge via the resulted multi-relational embedding.

\section*{Acknowledgements}

This paper was supported by the Early Career Scheme (ECS, No. 26206717) from Research Grants Council in Hong Kong and the Tencent AI Lab Rhino-Bird Focused Research Program.

% \clearpage

\bibliography{multiplex_word_embedding}

\begin{thebibliography}{35}
\expandafter\ifx\csname natexlab\endcsname\relax\def\natexlab#1{#1}\fi

\bibitem[{Bengio et~al.(2003)Bengio, Ducharme, Vincent, and
  Janvin}]{bengio2003-JMLR}
Yoshua Bengio, R{\'e}jean Ducharme, Pascal Vincent, and Christian Janvin. 2003.
\newblock A neural probabilistic language model.
\newblock \emph{Journal of Machine Learning Research}, 3:1137--1155.

\bibitem[{Bezdek and Hathaway(2003)}]{bezdek2003convergence}
James~C Bezdek and Richard~J Hathaway. 2003.
\newblock Convergence of alternating optimization.
\newblock \emph{Neural, Parallel \& Scientific Computations}, 11(4):351--368.

\bibitem[{Chen and Manning(2014)}]{chen2014fast}
Danqi Chen and Christopher Manning. 2014.
\newblock A fast and accurate dependency parser using neural networks.
\newblock In \emph{Proceedings of EMNLP}, pages 740--750.

\bibitem[{Collobert et~al.(2011)Collobert, Weston, Bottou, Karlen, Kavukcuoglu,
  and Kuksa}]{collobert2011-JMLR1}
Ronan Collobert, Jason Weston, L{\'e}on Bottou, Michael Karlen, Koray
  Kavukcuoglu, and Pavel Kuksa. 2011.
\newblock {Natural Language Processing (Almost) from Scratch}.
\newblock \emph{Journal of Machine Learning Research}, 12:2493--2537.

\bibitem[{de~Cruys(2014)}]{DBLP:conf/emnlp/Cruys14}
Tim~Van de~Cruys. 2014.
\newblock A neural network approach to selectional preference acquisition.
\newblock In \emph{Proceedings of EMNLP}, pages 26--35.

\bibitem[{Devlin et~al.(2019)Devlin, Chang, Lee, and
  Toutanova}]{devlin2018bert}
Jacob Devlin, Ming{-}Wei Chang, Kenton Lee, and Kristina Toutanova. 2019.
\newblock {BERT:} pre-training of deep bidirectional transformers for language
  understanding.
\newblock In \emph{Proceedings of NAACL-HLT 2019}, pages 4171--4186.

\bibitem[{Erk et~al.(2010)Erk, Pad{\'{o}}, and
  Pad{\'{o}}}]{DBLP:journals/coling/ErkPP10}
Katrin Erk, Sebastian Pad{\'{o}}, and Ulrike Pad{\'{o}}. 2010.
\newblock A flexible, corpus-driven model of regular and inverse selectional
  preferences.
\newblock \emph{Computational Linguistics}, 36(4):723--763.

\bibitem[{Hill et~al.(2015)Hill, Reichart, and
  Korhonen}]{DBLP:journals/coling/HillRK15}
Felix Hill, Roi Reichart, and Anna Korhonen. 2015.
\newblock Simlex-999: Evaluating semantic models with (genuine) similarity
  estimation.
\newblock \emph{Computational Linguistics}, 41(4):665--695.

\bibitem[{Hobbs(1978)}]{hobbs1978resolving}
Jerry~R Hobbs. 1978.
\newblock Resolving pronoun references.
\newblock \emph{Lingua}, 44(4):311--338.

\bibitem[{Inoue et~al.(2016)Inoue, Matsubayashi, Ono, Okazaki, and
  Inui}]{DBLP:conf/coling/InoueMOOI16}
Naoya Inoue, Yuichiroh Matsubayashi, Masayuki Ono, Naoaki Okazaki, and Kentaro
  Inui. 2016.
\newblock Modeling context-sensitive selectional preference with distributed
  representations.
\newblock In \emph{Proceedings of COLING}, pages 2829--2838.

\bibitem[{Keller and Lapata(2003)}]{DBLP:journals/coling/KellerL03}
Frank Keller and Mirella Lapata. 2003.
\newblock Using the web to obtain frequencies for unseen bigrams.
\newblock \emph{Computational Linguistics}, 29(3):459--484.

\bibitem[{Lee et~al.(2018)Lee, He, and Zettlemoyer}]{lee2018higher}
Kenton Lee, Luheng He, and Luke Zettlemoyer. 2018.
\newblock Higher-order coreference resolution with coarse-to-fine inference.
\newblock In \emph{Proceedings of NAACL-HLT}, pages 687--692.

\bibitem[{Levy and Goldberg(2014)}]{DBLP:conf/acl/LevyG14}
Omer Levy and Yoav Goldberg. 2014.
\newblock Dependency-based word embeddings.
\newblock In \emph{Proceedings of ACL}, pages 302--308.

\bibitem[{McRae et~al.(1998)McRae, Spivey-Knowlton, and
  Tanenhaus}]{mcrae1998modeling}
Ken McRae, Michael~J Spivey-Knowlton, and Michael~K Tanenhaus. 1998.
\newblock Modeling the influence of thematic fit (and other constraints) in
  on-line sentence comprehension.
\newblock \emph{Journal of Memory and Language}, 38(3):283--312.

\bibitem[{Mikolov et~al.(2013)Mikolov, Sutskever, Chen, Corrado, and
  Dean}]{DBLP:conf/nips/MikolovSCCD13}
Tomas Mikolov, Ilya Sutskever, Kai Chen, Gregory~S. Corrado, and Jeffrey Dean.
  2013.
\newblock Distributed representations of words and phrases and their
  compositionality.
\newblock In \emph{Proceedings of NIPS}, pages 3111--3119.

\bibitem[{Pennington et~al.(2014)Pennington, Socher, and
  Manning}]{DBLP:conf/emnlp/PenningtonSM14}
Jeffrey Pennington, Richard Socher, and Christopher~D. Manning. 2014.
\newblock Glove: Global vectors for word representation.
\newblock In \emph{Proceedings of EMNLP}, pages 1532--1543.

\bibitem[{Peters et~al.(2018)Peters, Neumann, Iyyer, Gardner, Clark, Lee, and
  Zettlemoyer}]{DBLP:conf/naacl/PetersNIGCLZ18}
Matthew~E. Peters, Mark Neumann, Mohit Iyyer, Matt Gardner, Christopher Clark,
  Kenton Lee, and Luke Zettlemoyer. 2018.
\newblock Deep contextualized word representations.
\newblock In \emph{Proceedings of NAACL-HLT}, pages 2227--2237.

\bibitem[{Resnik(1997)}]{resnik1997selectional}
Philip Resnik. 1997.
\newblock Selectional preference and sense disambiguation.
\newblock \emph{Tagging Text with Lexical Semantics: Why, What, and How?}

\bibitem[{Ritter et~al.(2010)Ritter, Mausam, and
  Etzioni}]{DBLP:conf/acl/RitterME10}
Alan Ritter, Mausam, and Oren Etzioni. 2010.
\newblock A latent dirichlet allocation method for selectional preferences.
\newblock In \emph{Proceedings of ACL}, pages 424--434.

\bibitem[{Rooth et~al.(1999)Rooth, Riezler, Prescher, Carroll, and
  Beil}]{rooth1999inducing}
Mats Rooth, Stefan Riezler, Detlef Prescher, Glenn Carroll, and Franz Beil.
  1999.
\newblock Inducing a semantically annotated lexicon via em-based clustering.
\newblock In \emph{Proceedings of ACL}, pages 104--111.

\bibitem[{Santus et~al.(2017)Santus, Chersoni, Lenci, and
  Blache}]{santus2017measuring}
Enrico Santus, Emmanuele Chersoni, Alessandro Lenci, and Philippe Blache. 2017.
\newblock Measuring thematic fit with distributional feature overlap.
\newblock In \emph{Proceedings of EMNLP}, pages 648--658.

\bibitem[{Solovyev et~al.(2017)Solovyev, Bochkarev, and
  Shevlyakova}]{solovyev2017dynamics}
Valery~D Solovyev, Vladimir~V Bochkarev, and Anna~V Shevlyakova. 2017.
\newblock Dynamics of core of language vocabulary.
\newblock \emph{arXiv preprint:1705.10112}.

\bibitem[{Song et~al.(2017)Song, Lee, and Xia}]{song-etal-2017-learning}
Yan Song, Chia-Jung Lee, and Fei Xia. 2017.
\newblock Learning word representations with regularization from prior
  knowledge.
\newblock In \emph{Proceedings of CoNLL}, pages 143--152.

\bibitem[{Song and Shi(2018)}]{ijcai2018-607}
Yan Song and Shuming Shi. 2018.
\newblock Complementary learning of word embeddings.
\newblock In \emph{Proceedings of IJCAI 2018}, pages 4368--4374.

\bibitem[{Song et~al.(2018{\natexlab{a}})Song, Shi, and Li}]{ijcai2018-0608}
Yan Song, Shuming Shi, and Jing Li. 2018{\natexlab{a}}.
\newblock Joint learning embeddings for chinese words and their components via
  ladder structured networks.
\newblock In \emph{Proceedings of IJCAI 2018}, pages 4375--4381.

\bibitem[{Song et~al.(2018{\natexlab{b}})Song, Shi, Li, and
  Zhang}]{song-etal-2018-directional}
Yan Song, Shuming Shi, Jing Li, and Haisong Zhang. 2018{\natexlab{b}}.
\newblock Directional skip-gram: Explicitly distinguishing left and right
  context for word embeddings.
\newblock In \emph{Proceedings of NAACL-HLT}, pages 175--180.

\bibitem[{Tang et~al.(2016)Tang, Xiong, Zhang, and
  Gong}]{DBLP:conf/coling/TangXZG16}
Haiqing Tang, Deyi Xiong, Min Zhang, and Zhengxian Gong. 2016.
\newblock Improving statistical machine translation with selectional
  preferences.
\newblock In \emph{Proceedings of COLING}, pages 2154--2163.

\bibitem[{Turney and Pantel(2010)}]{turney2010-JAIR}
Peter~D. Turney and Patrick Pantel. 2010.
\newblock {From Frequency to Meaning: Vector Space Models of Semantics}.
\newblock \emph{{Journal of Artificial Intelligence Research}}, 37(1):141--188.

\bibitem[{Wilks(1975)}]{wilks1975preferential}
Yorick Wilks. 1975.
\newblock A preferential, pattern-seeking, semantics for natural language
  inference.
\newblock \emph{Artificial intelligence}, 6(1):53--74.

\bibitem[{Zapirain et~al.(2013)Zapirain, Agirre, M{\`{a}}rquez, and
  Surdeanu}]{semantic_role_classification}
Be{\~{n}}at Zapirain, Eneko Agirre, Llu{\'{\i}}s M{\`{a}}rquez, and Mihai
  Surdeanu. 2013.
\newblock Selectional preferences for semantic role classification.
\newblock \emph{Computational Linguistics}, 39(3):631--663.

\bibitem[{Zhang et~al.(2019{\natexlab{a}})Zhang, Ding, and
  Song}]{DBLP:conf/acl/ZhangDS19}
Hongming Zhang, Hantian Ding, and Yangqiu Song. 2019{\natexlab{a}}.
\newblock {SP-10K:} {A} large-scale evaluation set for selectional preference
  acquisition.
\newblock In \emph{Proceedings of ACL 2019}, pages 722--731.

\bibitem[{Zhang et~al.(2018)Zhang, Qiu, Yi, and
  Song}]{DBLP:conf/ijcai/ZhangQYS18}
Hongming Zhang, Liwei Qiu, Lingling Yi, and Yangqiu Song. 2018.
\newblock Scalable multiplex network embedding.
\newblock In \emph{Proceedings of IJCAI 2019}, pages 3082--3088.

\bibitem[{Zhang et~al.(2019{\natexlab{b}})Zhang, Song, and
  Song}]{DBLP:conf/naacl/ZhangSS19}
Hongming Zhang, Yan Song, and Yangqiu Song. 2019{\natexlab{b}}.
\newblock Incorporating context and external knowledge for pronoun coreference
  resolution.
\newblock In \emph{Proceedings of NAACL-HLT 2019}, pages 872--881.

\bibitem[{Zhang et~al.(2019{\natexlab{c}})Zhang, Song, Song, and
  Yu}]{DBLP:conf/acl/ZhangSSY19}
Hongming Zhang, Yan Song, Yangqiu Song, and Dong Yu. 2019{\natexlab{c}}.
\newblock Knowledge-aware pronoun coreference resolution.
\newblock In \emph{Proceedings of ACL 2019}, pages 867--876.

\bibitem[{Zou et~al.(2013)Zou, Socher, Cer, and Manning}]{zou2013bilingual}
Will~Y Zou, Richard Socher, Daniel Cer, and Christopher~D Manning. 2013.
\newblock Bilingual word embeddings for phrase-based machine translation.
\newblock In \emph{Proceedings of EMNLP}, pages 1393--1398.

\end{thebibliography}
\bibliographystyle{acl_natbib}

\end{document}